\documentclass{hld2025} 

\usepackage[utf8]{inputenc} 
\usepackage[T1]{fontenc}    
\usepackage{hyperref}       
\usepackage{url}            
\usepackage{booktabs}       
\usepackage{amsfonts}       
\usepackage{nicefrac}       
\usepackage{microtype}      
\usepackage{xcolor}         
\usepackage{amsmath}
\usepackage{amssymb}
\usepackage{mathtools}
\usepackage{bbm}
\usepackage{graphicx}
\usepackage{algorithm}
\usepackage{algorithm2e}

\usepackage{amsmath,amsfonts,bm}
\usepackage{dsfont}

\def\eqref#1{equation~\ref{#1}}

\DeclareMathAlphabet{\mathsfit}{\encodingdefault}{\sfdefault}{m}{sl}
\SetMathAlphabet{\mathsfit}{bold}{\encodingdefault}{\sfdefault}{bx}{n}

\newcommand{\Var}{\mathrm{Var}}

\usepackage[normalem]{ulem}

\usepackage{natbib}

\title{On the Performance of Differentially Private Optimization with Heavy-Tail Class Imbalance}

\hldauthor{%
\Name{Qiaoyue Tang} \Email{qiaoyuet@cs.ubc.ca}\\
\addr {University of British Columbia, Vancouver, Canada} 
\AND
\Name{Alain Zhiyanov} \Email{alainzhiyanov@gmail.com}\\
\addr {University of British Columbia, Vancouver, Canada} 
\AND
\Name{Mathias Lécuyer} \Email{mathias.lecuyer@ubc.ca}\\
\addr {University of British Columbia, Vancouver, Canada} 
}

\begin{document}

\maketitle
\vspace{-1cm}

\section{Introduction}
The recent success of large language models has unlocked remarkable advancements in many real-world prediction and generation tasks. However, their reliance on training with massive text corpora, potentially including data collected from unknown sources or without proper authorization, raises significant concerns regarding privacy risks of exposing unintended information \citep{sun2024doesfinetuninggpt3openai, carlini2021extractingtrainingdatalarge}. Such privacy risk has been shown to emerge in model pre-training, as well as when fine-tuning on downstream tasks \citep{mireshghallah2022memorizationnlpfinetuningmethods}. Differential Privacy (DP) is an effective privacy mechanism with formal and rigorous guarantee, commonly incorporated into the model training process to establish upper-bounds on information leakage from the training data. However, training language models with DP often results in substantial performance degradation. In this work, we build upon two recent findings---the ill-conditioning of optimization under heavy-tail class imbalance, and bias in the Adam optimizer introduced by DP noise---to analyze the performance of privately learning with heavy-tail imbalanced labels, a common setting in next-token-prediction tasks used to learn language models.

\paragraph{Performance Gap between SGD and Adam in Language Models.}
The empirical observation that Adam often outperforms Stochastic Gradient Descent (SGD) on language tasks, in contrast to the other tasks, has motivated many studies investigating Adam's optimization behaviour in language models. \cite{liu2023understandingdifficultytrainingtransformers} shows that Adam updates parameters uniformly despite potentially larger difference in gradient scales in transformer models. Following the line of work, \cite{zhang2020adaptivemethodsgoodattention} attributes the performance gap to gradient noise in language transformers compared to vision tasks, and shows that Adam exhibits greater robustness to heavy-tailed noise in stochastic gradients. Subsequently, \cite{kunstner2023noisemainfactorgap} shows that the performance gap persists even in the full-batch training setting and differentiates Adam from SGD by relating Adam's behaviour to sign descent.
Recently, \cite{kunstner2024heavytailedclassimbalanceadam} investigates the performance gap from a novel perspective of heavy-tail class imbalance. They show that the gradient and Hessian with respect to model parameters for each class are both dominated by the relative class frequency. As a consequence, they show in the scenario which the classes comprise a few high-frequency classes and many low-frequency classes, Adam benefits from estimating the curvature and could learn low-frequency classes faster than Gradient Descent (GD) while maintaining comparable performance for the high-frequency classes.

\paragraph{Bias Correction in Differentially Private Learning with Adam.}
One common approach to train modern deep learning models with privacy guarantee is to integrate DP guarantee with (Stochastic) Gradient Descent, i.e. DP-(S)GD \citep{Abadi_2016}. The key steps in DP-(S)GD are to (1) clip per-sample gradients with $\ell_2$-clipping-threshold $C$ such that the change in model output when changing the model input is well-bounded, and (2) add isotropic DP noise randomly drawn from a $\mathcal{N}(0, \sigma^2C^2)$ Gaussian distribution to the aggregated clipped gradients. With the same `recipe' to privatize gradient, previous works have also use DP-Adam, which replaces the gradient in the Adam optimizer with a private gradient.
While a performance gap between Adam and SGD in language tasks is consistently observed in non-private learning, this difference appears to be smaller in private learning scenarios. \cite{tang2023dpadambcdpadamactuallydpsgd} shows that DP-Adam behaves like DP-SGD (with momentum) since the second moment estimates in Adam's update is largely dominated by the variance of DP noise. As Adam uses second moment as an estimate of curvature to normalize the gradient, the dominating additive bias caused by DP noise potentially makes the estimation ineffective. The authors show that by removing the bias, the corrected algorithm (DP-AdamBC) more closely resembles sign descent---similar to Adam's behaviour in non-private learning---and empirically improves classification performance in several tasks. Appendix \ref{appendix:pseudo-code} includes the pseudo-code for the three DP optimization algorithms.

\paragraph{Contributions.}
In this work, we analyze the optimization behaviour of common private learning optimization algorithms under heavy-tail class imbalanced distribution. We show that, in a stylized model, optimizing with Gradient Descent with differential privacy (DP-GD) suffers when learning low-frequency classes, whereas optimization algorithms that estimate second-order information do not  (\S\ref{sec:linear_model}). In particular, DP-AdamBC that removes the DP bias from estimating loss curvature is a crucial component to avoid the ill-condition caused by heavy-tail class imbalance, and empirically fits the data better with $\approx8\%$ and $\approx5\%$ increase in training accuracy when learning the least frequent classes on both controlled experiments (\S\ref{subsec:experiment_linear}) and real data (Appendix \ref{appendix:exp_real_data}) respectively.

\section{An Analysis on Linear Models}
\label{sec:linear_model}
We use the same linear model setup as \cite{kunstner2024heavytailedclassimbalanceadam} to mathematically demonstrate the difficulty of learning heavy-tail imbalanced datasets with differential privacy, and the advantage of bias correction in Adam to relieve the ill-conditioning due to class imbalance. 
%
The inputs $\mathbf{X}$ are generated from a $d$-dimensional uniform distribution on $[0, 1]^d$. The classes $\mathbf{y}$ are generated to approximately follow the Zipf distribution, in which the $k$-th frequent class has frequency $\propto 1/k$, $k \in [1, \ldots,c]$. We train a linear model with weights $\mathbf{W}\in \mathbb{R}^{c\times d}$ to minimize the loss $\ell(\mathbf{W}, \mathbf{x},y)=-\log(s(\mathbf{Wx})_y)$ where $s(z)_k=e^{z_k}/\sum_j e^{z_j}$ is the softmax function. The optimization objective is to minimize the mean loss over $n$ training samples $\mathcal{L}(\mathbf{W,x,y})=\frac{1}{n}\sum_{i=1}^{n} \ell(\mathbf{W}, \mathbf{x}_i,y_i)$.
The linear model setup allows analyzing the gradient vector and Hessian matrix with respect to each row $\mathbf{w}_1, \ldots, \mathbf{w}_c$ of the weight matrix $\mathbf{W}$, which is used to illustrate ill-condition from heavy-tail class imbalance in non-private learning (\S\ref{subsec:ill-condition-background}). We use the mathematical expressions of private gradient and Hessian to demonstrate the ill-condition under private learning in \S\ref{subsec:dp-noise-ill-condition},\ref{subsec:clipping-ill-condition} and demonstrate the empirical experiments on such setup with synthetic data in \S\ref{subsec:experiment_linear}.
%
We refer to \cite{kunstner2024heavytailedclassimbalanceadam} for the complete detail of derivations.

\paragraph{Hessian Ill-condition from Heavy-tail Imbalance in Non-private Learning.}
\label{subsec:ill-condition-background}
In an optimization problem, the gradient indicates the direction of fastest local change in the loss function, whereas the Hessian indicates the curvature of the loss function, i.e. the rate of change of the gradient with respect to each parameter. Intuitively, optimization is more difficult if the loss curvature differs significantly along different directions: following the gradient vector alone could bounce back and forth in high curvature directions and make slow progress in low curvature directions \citep{JMLR:v12:duchi11a, yang2023sidescoinlimitsuntuned, anil2021scalablesecondorderoptimization}. Since the learning rate scales gradients equally in all directions, adapting this hyper-parameter cannot alleviate such curvature challenges.
%

\citet{kunstner2024heavytailedclassimbalanceadam} demonstrate that,
under a heavy-tail imbalanced class distribution where most samples come from low-frequency classes, the learning problem is ill-conditioned.
Indeed, under the assumption that the model correctly assigns data samples to class $k$ with a large output probability (logit) $p$ (Assumption 1, \cite{kunstner2024heavytailedclassimbalanceadam}),
the gradient and Hessian with respect to model parameters of each class are dominated by the relative class frequency $\pi_k=n_k/n$.
Specifically, \citet{kunstner2024heavytailedclassimbalanceadam} show that in that case, $\mathbf{g}_t \coloneqq\nabla_\mathbf{w_k} \mathcal{L} = (1-p)\pi_k \bar{\mathbf{x}}^k + O(1/c)$, and $\mathbf{H}_t \coloneqq \nabla^2_\mathbf{w_k} \mathcal{L}= p(1-p)\pi_k \bar{\mathbf{H}}^k + O(1/c)$, where $\bar{\mathbf{x}}^k=(1/n_k)\sum_{i=1|y_i=k}^{n_k}\mathbf{x}_i$, $\bar{\mathbf{H}}^k=(1/n_k)\sum_{i=1|y_i=k}^{n_k}\mathbf{x}_i\mathbf{x}_i^\top$. 

Gradient Descent does not account for the underlying loss curvature and only updates the parameters following the gradient.
Under the above class imbalance model, we can see that the gradient is small for low-frequency classes (small $\pi_k$): this leads to slow update in low-curvature regions since the Hessian values are also small. At the opposite, the gradient is large for high-frequency classes (large $\pi_k$), resulting in large updates in high-curvature directions, as the Hessian is also large. As a result, Gradient Descent suffers from slow convergence in learning low-frequency classes, while increasing the learning rate makes updates to high-frequency classes unstable (the parameters oscillates around good regions).

A natural fix to the ill-conditioning is to take into account the curvature information \citep{JMLR:v12:duchi11a}, using second-order methods that update parameters with $\Delta = \mathbf{g}_t\mathbf{H}_t^{-1}$. Since calculating $\mathbf{H}_t^{-1}$ is computationally expensive, and can be unstable, Adam uses the inverse of the gradients' second moment as a proxy to the diagonal of Hessian with cheap computation cost, which adapts gradients to the curvature and alleviates the effect of ill-conditioning \citep{kingma2017adammethodstochasticoptimization, JMLR:v12:duchi11a, martens2020optimizingneuralnetworkskroneckerfactored}.

\paragraph{Effect of DP Noise with Heavy-tail Imbalance.}
\label{subsec:dp-noise-ill-condition}
We first assume that no gradients are clipped, which is empirically achievable by setting a large $\ell_2$-clipping threshold such that $\forall i, C \geq ||\mathbf{g}_t(\mathbf{x}_i)||$. We discuss the effect of gradient clipping in the next section.
%
Let $\sigma$ be the noise multiplier that controls the privacy level: the private gradient $\Tilde{\mathbf{g}}_t$ is obtained by adding DP noise sampled from the Gaussian distribution $\mathbf{z_t} \sim \mathcal{N}(0, \sigma^2C^2\mathbb{I})$. For the linear model defined above, we have $\widetilde{\mathbf{g}}_t = \nabla_\mathbf{w_k} \mathcal{L} + \mathbf{z_t} \approx (1-p)\pi_k \bar{\mathbf{x}}^k + \mathbf{z_t}$. 
As before, the Hessian is $\mathbf{H}_t=\nabla^2_\mathbf{w_k} \mathcal{L} \approx p(1-p)\pi_k \bar{\mathbf{H}}^k$. It is usually not explicitly calculated in DP optimization algorithms, but it reflects the true loss curvature with respect to the model parameters at step $t$. In the following, we compare the scale of the private gradient and the Hessian matrix with respect to the rows $\mathbf{w}_1, \ldots, \mathbf{w}_c$ of the weight matrix $\mathbf{W}$, for different DP optimization algorithms.

\textbf{DP-GD} updates parameters using the private gradient only: $\Delta^{\textnormal{DP-GD}} = \widetilde{\mathbf{g}}_t$. Since the DP noise distribution is mean zero, the private gradient is the same as the non-private gradient in expectation, $\mathbb{E}_{Z}[\widetilde{\mathbf{g}}_t] = \nabla_\mathbf{w_k} \mathcal{L}$. Under heavy-tail imbalanced labels ill-conditioning persists, because on average DP-GD takes large steps for high-frequency classes in high-curvature regions (large $\mathbb{E}_{Z}[\widetilde{\mathbf{g}}_t]$ and large $\mathbf{H}_t$ from large $\pi_k$), and small steps for low-frequency classes in low-curvature regions (small $\mathbb{E}_{Z}[\widetilde{\mathbf{g}}_t]$ and small $\mathbf{H}_t$ from small $\pi_k$). Since the model parameters are updated with the same scalar learning rate $\eta_t$ across the whole private gradient vector, $\theta_{t+1} \xleftarrow{} \theta_t - \eta_t \widetilde{\mathbf{g}}_t$, DP-GD is either slow in learning the low-frequency classes or unstable in learning the high-frequency classes.

Adam relieves the ill-conditioning in non-private learning. However, under additive DP noise, achieving a similar effect requires applying  bias correction in estimating curvature with noisy gradients.
Adam preconditions the gradient by (the square root of) the moving average of squared gradient $v_t = \sum_{\tau=1}^{t}\beta^{t-\tau}\mathbf{g}_{\tau}^2$, \footnote{The constant scaling of $1-\beta$ applies to all the estimates and is ignored here. We also exclude the numerical stability constants in Adam's update in both DP-Adam and DP-AdamBC.} which approximates the second moment of the gradient in expectation, $\mathbb{E}[v_t] \approx \mathbb{E}[\mathbf{g}_t^2]$ \citep{kingma2017adammethodstochasticoptimization}. With the privacy constraint, \textbf{DP-Adam} estimates the curvature from the noisy gradient using $\widetilde{v}_t =\sum_{\tau=1}^{t}\beta^{t-\tau}\widetilde{\mathbf{g}}_{\tau}^2$, which biases the estimation as $\mathbb{E}[\widetilde{v}_t] \approx \mathbb{E}[\widetilde{\mathbf{g}}_t^2] = \mathbb{E}[\mathbf{g}_t^2]+\Var[\mathbf{z_t}]$. 
\cite{tang2023dpadambcdpadamactuallydpsgd} shows that empirically $\Var[\mathbf{z_t}]$ often dominates $\mathbb{E}[\mathbf{g}_t^2]$. In that case, DP noise makes the second moment estimation inaccurate and lead to incorrectly scaling the gradient, thus failing to alleviate ill-conditioning. In the linear model example, as $\mathbb{E}_{Z}[\widetilde{\mathbf{g}}_t] \approx (1-p)\pi_k\bar{\mathbf{x}}^k$, the curvature estimate from DP-Adam is $\mathbb{E}[\widetilde{v}_t] \approx ((1-p)\pi_k\bar{\mathbf{x}}^k)^2 + \sigma^2C^2$. When the DP noise is large, either from a large $\sigma$ to ensure higher privacy guarantee or a large $C$ for better empirical performance, $\mathbb{E}[\widetilde{v}_t]$ cannot effectively rescale $\mathbb{E}_{Z}[\widetilde{\mathbf{g}}_t]$ as expected. For low/high-frequency classes with small/large $\mathbb{E}_{Z}[\widetilde{\mathbf{g}}_t]$ values from small/large $\pi_k$ respectively, the gradients are scaled similarly with $\Var[\mathbf{z_t}]$. Therefore, similar to DP-GD, the low/high-frequency classes continue to update with small/large steps.

\textbf{DP-AdamBC} corrects the bias by replacing $\widetilde{v}_t$ by $\widetilde{v}_t - \Var[\mathbf{z_t}]$, which restores $\mathbb{E}[\widetilde{v}_t] \approx \mathbb{E}[\mathbf{g}_t^2]$  better adapting gradients with respect to unbiased estimates of curvature. For low/high-frequency classes with small/large $\mathbb{E}_{Z}[\widetilde{\mathbf{g}}_t]$ values, scaling by $\mathbb{E}[\mathbf{g}_t^2]$ magnifies/shrinks the update for low/high-frequency classes by dividing a small/large value of $\mathbb{E}[\mathbf{g}_t^2]$, respectively. Therefore, as scaling gradient by the curvature relieves the heavy-tail imbalanced ill-conditioning in the non-private case, scaling the private gradient by the curvature estimator with DP bias removed better relieves the ill-conditioning in private learning.

\begin{figure}[H]
    \centering
    \includegraphics[width=0.85\textwidth]{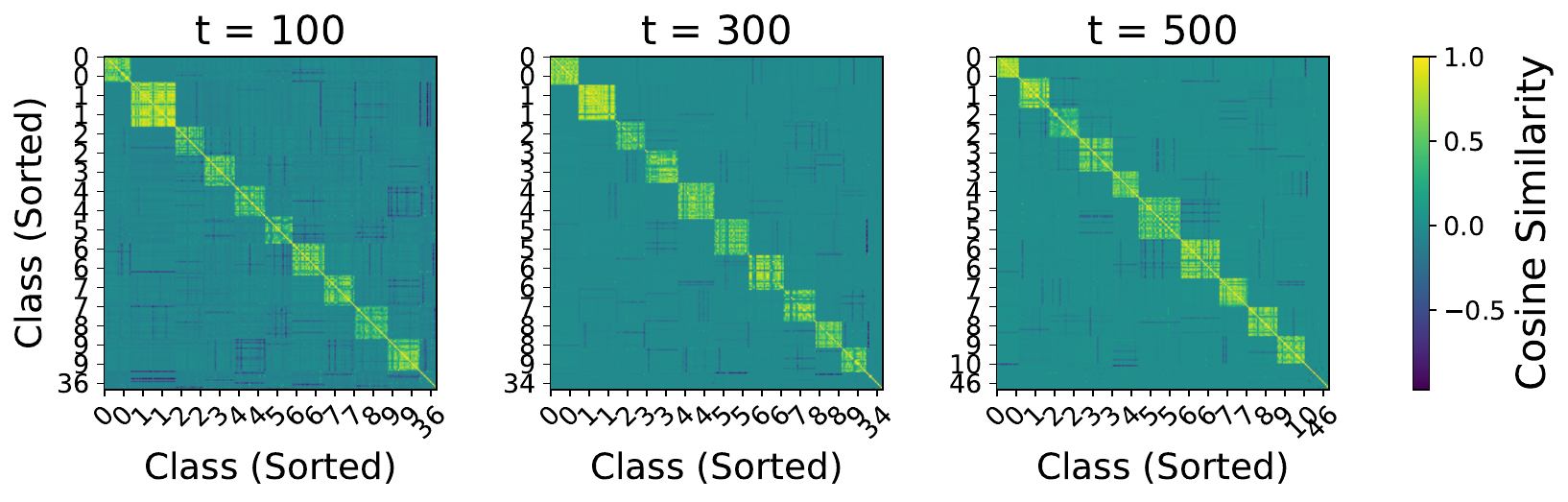}
    \caption{The pairwise cosine similarity between (520 randomly sampled) per-sample gradient at different training steps. The gradients are more aligned within the same class with higher cosine similarity values and less aligned across classes with lower values.}
    \label{fig:cosine_sim_linear}
\end{figure}

\paragraph{Effect of Gradient Clipping with Heavy-tail Imbalance.}
\label{subsec:clipping-ill-condition}
Gradient clipping is essential for providing a valid privacy guarantee in neural networks. 
One might have thought that it would also partially alleviate the effect of ill-conditioning by dampening the gradient's dependency on class frequency (large gradients for high frequency classes would get clipped more). And indeed, in non-private learning cases \citet{kunstner2024heavytailedclassimbalanceadam} notice that normalized Gradient Descent, i.e. update parameters with $\Delta=\mathbf{g_t}/||\mathbf{g_t}||_2$, can improve performance in heavy tailed class imbalanced datasets.
DP gradient clipping, however, clips the gradients of individual data points before aggregation: $\widetilde{\mathbf{g}}_t = (1/n)(\sum_{i=1}^{n}\textnormal{clip}(\nabla_{\mathbf{w_k}} \mathcal{L}(\mathbf{x}_i), \, C) + \mathbf{z}_t)$, where $\textnormal{clip}(\mathbf{g}_t(\mathbf{x_i}), \,C) = \mathbf{g}_t(\mathbf{x_i})/\max{(1, ||\mathbf{g}_t(\mathbf{x_i})||_2/C)}$ clips per-sample gradient with $||\mathbf{g}||_2\geq C$ to have $\ell_2$-norm equals $C$. 
For each clipping per-sample gradient, the magnitude decreases ($(\mathbf{g}_t(\mathbf{x_i})/||\mathbf{g}_t(\mathbf{x_i})||_2)\cdot C$), but the direction is preserved. The gradient's component towards the correct class $k$ (the values of the gradient in $\mathbf{w}_k$) remains the largest, and accumulates over data points proportionally to the class frequency $\pi_k$.
The problem is hence still ill-conditioned, and changing the learning rate is not sufficient: one needs to account for curvature, as Adam does.
%
Empirically, we can examine the cosine-similarity among per-sample gradient pairs, to confirm that gradients for the same class are more aligned (smaller angle), and will thus accumulate proportionally to $\pi_k$ despite clipping. Figure \ref{fig:cosine_sim_linear} shows the cosine similarities for the gradients with respect to the weight matrix $\mathbf{W}$ in the linear model example. We observe that the values are higher among samples within the same class than across different classes, which means the gradient of examples within the same class are more aligned, whereas the gradients of examples from different classes tend to be orthogonal to each other. Since gradient clipping in private learning re-aggregates clipped per-sample gradients, we could rewrite the expression and split by classes, $(1/n)\sum_{i=1}^{n}\textnormal{clip}(\mathbf{g}_t(\mathbf{x_i}), \, C) = (1/n)\sum_{k=1}^{c}\sum_{i|y_i=k}\textnormal{clip}(\mathbf{g}_t(\mathbf{x_i}), \, C) = \sum_{k=1}^{c}(\pi_k/n_k)\sum_{i|y_i=k}\textnormal{clip}(\mathbf{g}_t(\mathbf{x_i}), \, C)=\sum_{k=1}^{c}\pi_k\frac{\sum_{i|y_i=k}\textnormal{clip}((\mathbf{g}_t(\mathbf{x_i}), \, C)}{n_k}$. As a consequence, as per-sample gradients within the same class are more aligned, the aggregated clipped gradient before adding DP noise also has a similar dependence on relative class frequency $\pi_k$.

\section{Experiments on Linear Model with Synthetic Data}
\label{subsec:experiment_linear}
To examine the claims above, we generate synthetic data that exhibit heavy-tail class imbalance which has $m$ groups of classes with class sizes equals $c=2^{m+1}-1$. We set a minimum class size equals $5$ since extremely rare classes are difficult to learn under empirically meaningful DP guarantees. The inputs are drawn independent of the class labels with $n=m2^m$ examples in $d=2^m+n$ dimensions. We run under the full batch setting (batch size = $n$) to eliminate sampling bias with class imbalance. We show the result with $C=1$ where all per-sample gradients are clipped and $\sigma=10$ as the privacy level, with additional experiments included in Appendix \ref{appendix:vary_c_sigma}. We tune the learning rate of half a power ($1e^{-2}, 5e^{-2}, 1e^{-3}, ...$) for all optimizers and additionally tune the numerical stability constants in DP-Adam and DP-AdamBC. Since we focus on the training behaviour of the optimizers, we let it run until the overall training loss stops decreasing to observe the complete training trajectory, where the x-axis shows the epsilon value for the privacy guarantee over steps. We report the results of each optimizer with the lowest overall training loss. 

\begin{figure}[H]
    \centering
    \includegraphics[width=0.89\textwidth]{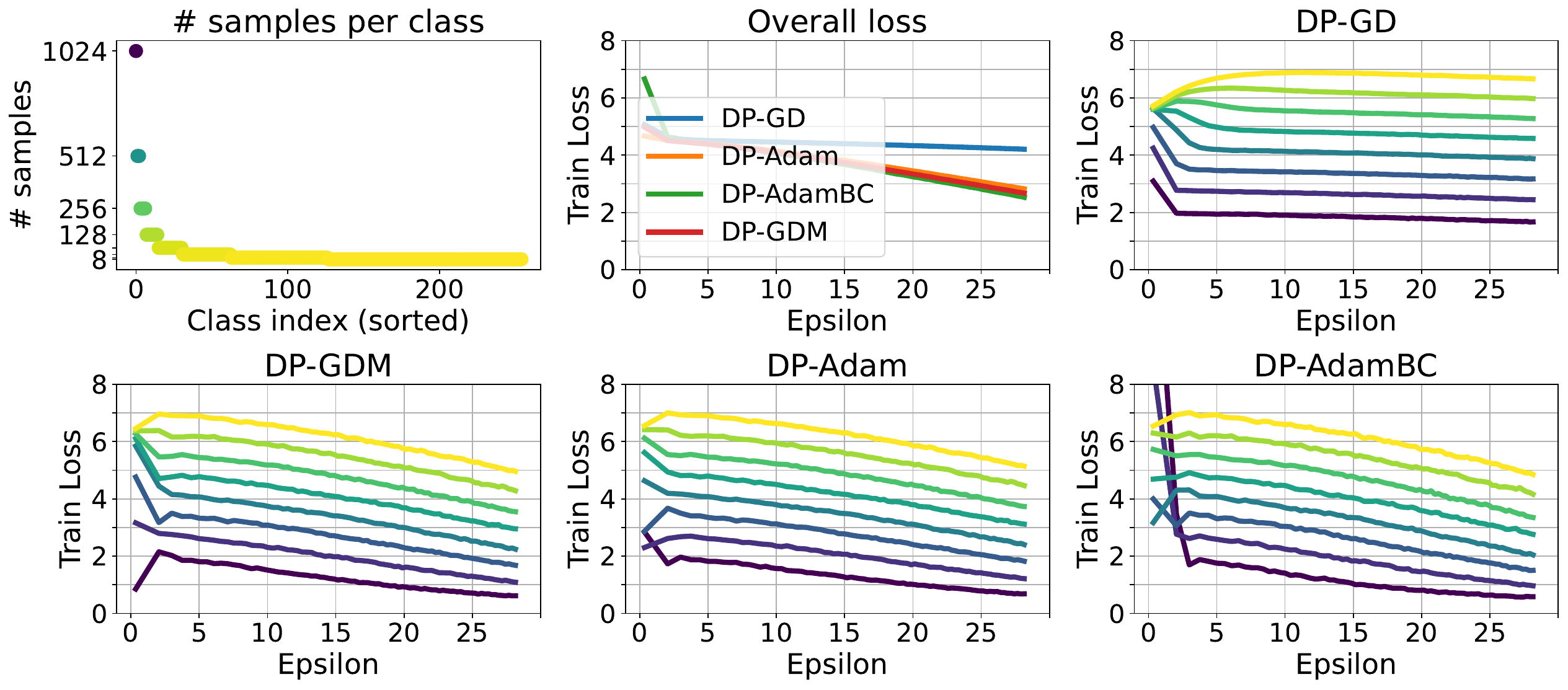}
    \includegraphics[width=0.89\textwidth]{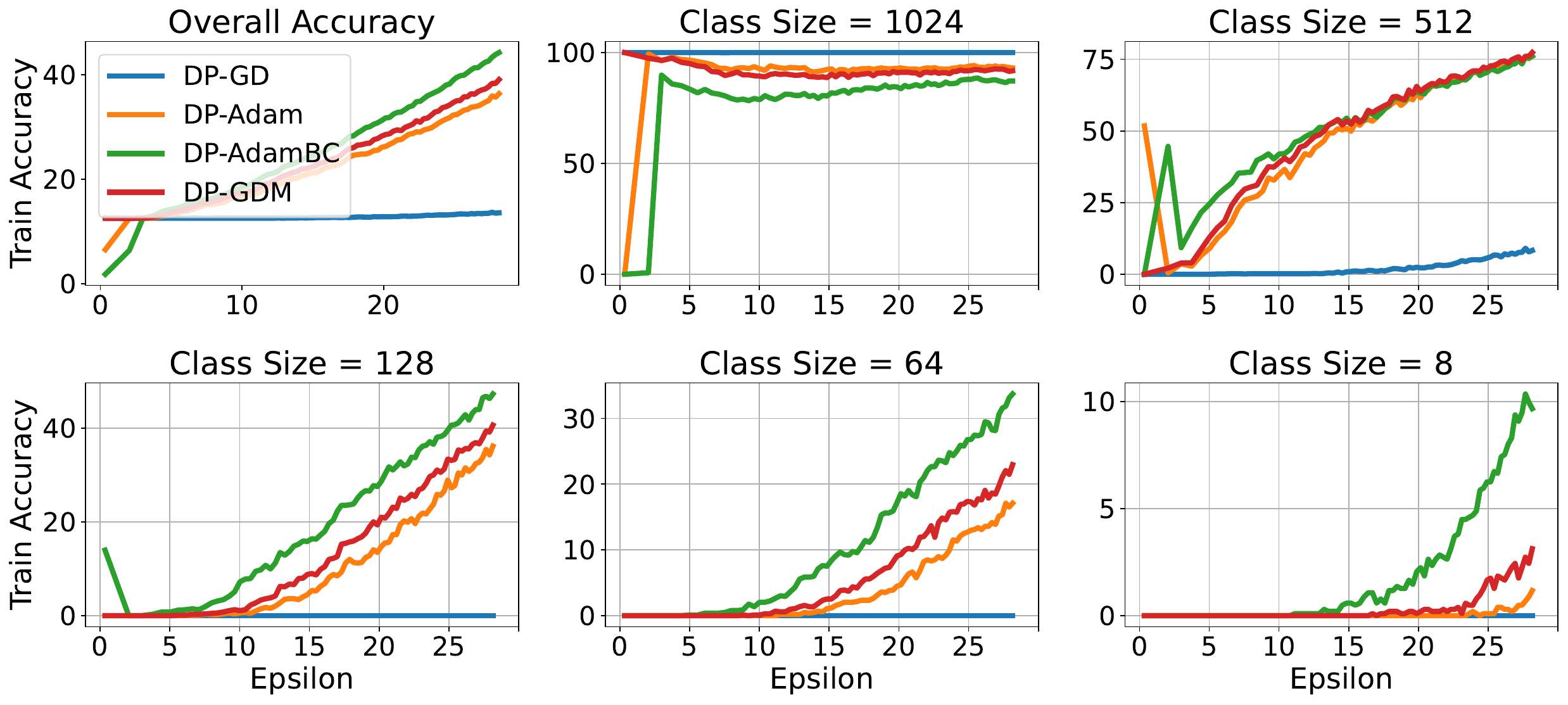}
    \caption{Results from synthetic dataset with linear model with $C=1, \sigma=10$. Subplots are ordered left to right, top to bottom (a)–(l). (a) The class distribution of the synthetic dataset. (b) The mean training loss from averaging all samples. (c)-(f) The mean training loss separated by averaging samples with the same label frequency. (g) The mean training accuracy from averaging all samples. (h)-(l) The mean training accuracy separated by averaging samples with the same label frequency.}
    \label{fig:c1n10_loss}
\end{figure}

Figure \ref{fig:c1n10_loss} shows the training loss and accuracy on the synthetic dataset with $n=8192, d=9216, c=255$, where the class distribution follows heavy-tail class imbalance as shown in Figure \ref{fig:c1n10_loss}(a). From the training losses, we observe that DP-GD has a flatter trends comparing to the others which has decreasing training losses especially for lower-frequency classes. For the lowest frequency group that has 128 classes with 8 samples in each class, the final training loss for the optimizers are DP-GD: 6.7, DP-GDM (DP-GD with Momentum): 4.9, DP-Adam:5.1 and DP-AdamBC: 4.8 respectively. We see that DP-GD fits the low-frequency classes the worst with the largest loss, whereas DP-AdamBC has the lowest loss even though the values among DP-GDM, DP-Adam and DP-AdamBC have smaller distinctions. From the training accuracies, we see that the gap in the overall accuracy is small when we aggregate samples over different classes, but the gap becomes larger for lower-frequency classes when separating samples by class sizes. For the highest frequency class with 1024 samples, DP-GD achieves 100.0\% training accuracy whereas the other optimizers have DP-GDM: 92.3\%, DP-Adam: 93.0\%, DP-AdamBC: 87.5\%. We observe that DP-GD performs better for high-frequency classes whereas the optimizers perform adequately well, whereas DP-AdamBC gains more advantage for lower-frequency classes. For the medium-frequency group with 8 classes and 128 samples each, DP-AdamBC has 47\% training accuracy while the other optimizers have DP-GD: 0\%, DP-GDM: 41\% and DP-Adam: 36\%. For the lowest frequency group with 128 classes and 8 samples each, DP-AdamBC performs the best with 9.5\%, which is approximately 9\%, 7\% and 8\% better than DP-GD, DP-GDM and DP-Adam respectively. In general, We observe that DP-GD learns the high frequency classes better than the other optimizers but cannot efficiently learn the low-frequency classes with comparable accuracies. Momentum (DP-GDM) that accumulates gradient over steps and pre-conditioning by a biased estimate of curvature (DP-Adam) helps relieve the situation, but DP-AdamBC has more balanced results as learning the high-frequency classes with comparable accuracy and gaining advantages in learning the lower-frequency classes.

\newpage

\section*{Acknowledgement}
We thank Erell Boutin Jeanniard du Dot for performing additional experimental work that contribute to this study. We are grateful for the support of the Natural Sciences and Engineering Research Council of Canada (NSERC) [reference number RGPIN-2022-04469]. This research was enabled by computational support provided by the Digital Research Alliance of Canada (alliancecan.ca), and by the University of British Columbia’s Advanced Research Computing (UBC ARC).

\bibliography{ref}

\newpage
\appendix

\section{Pseudo-code for Common DP Optimization Algorithms}
\label{appendix:pseudo-code}

\begin{algorithm}[h]
\caption{DP-GD and DP-GD with Momentum}
\KwIn{Dataset $\mathcal{D}$, loss function $\mathcal{L}(\theta, x)$, learning rate $\eta_t$, noise scale $\sigma$, clipping norm $C$, sample size $L$, number of iterations $T$, momentum parameter $\mu$}
Initialize model parameters $\theta_0$\;\\
\For{$t = 1$ to $T$}{
    \ForAll{$x_i$}{
        Compute gradient: $g_t(x_i) \gets \nabla_\theta \mathcal{L}(\theta_{t-1}, x_i)$\; \\
        Clip gradient: $\bar{g}_t(x_i) \gets \frac{g_t(x_i)}{\max\left(1, \frac{\|g_t(x_i)\|_2}{C}\right)}$
    }
    Aggregate clipped gradients: $\tilde{g}_t \gets \frac{1}{L} \sum_{i=1}^{L} \bar{g}_t(x_i) + \frac{1}{L}\mathcal{N}(0, \sigma^2 C^2 I)$\;\\
    Update parameters:
    \begin{equation*}
    \textnormal{(DP-GD)}\;\;\theta_t \gets \theta_{t-1} - \eta_t \tilde{g}_t
    \end{equation*}
\begin{equation*}
    \begin{gathered}
        \textnormal{(DP-GDM)}\;\;b_t \gets \begin{cases} 
          \tilde{g}_t & t=1 \\
          \mu b_{t-1} + \tilde{g}_t & t>1
        \end{cases} \\
        \theta_t \gets \theta_{t-1} - \eta_t b_t
    \end{gathered}
    \end{equation*}
}
\end{algorithm}

\begin{algorithm}[h]
\caption{DP-Adam and DP-AdamBC}
\KwIn{Dataset $\mathcal{D}$, loss function $\mathcal{L}(\theta, x)$, learning rate $\eta_t$, noise scale $\sigma$, clipping norm $C$, sample size $L$, number of iterations $T$, Adam hyperparameters $\beta_1$, $\beta_2$, $\gamma$, $\gamma'$}
Initialize parameters $\theta_0$, first moment vector $m_0 = 0$, second moment vector $v_0 = 0$\;\\
\For{$t = 1$ to $T$}{
    \ForAll{$x_i$}{
        Compute gradient: $g_t(x_i) \gets \nabla_\theta \mathcal{L}(\theta_{t-1}, x_i)$\;\\
        Clip gradient: $\bar{g}_t(x_i) \gets \frac{g_t(x_i)}{\max\left(1, \frac{\|g_t(x_i)\|_2}{C}\right)}$\;
    }
    Aggregate clipped gradients: $\tilde{g}_t \gets \frac{1}{L} \sum_{i=1}^{L} \bar{g}_t(x_i) + \frac{1}{L}\mathcal{N}(0, \sigma^2 C^2 I)$\;\\
    Update biased first moment estimate: $m_t \gets \beta_1 m_{t-1} + (1 - \beta_1)\tilde{g}_t$\;\\
    Update biased second moment estimate: $v_t \gets \beta_2 v_{t-1} + (1 - \beta_2)\tilde{g}_t^2$\;\\
    Compute bias-corrected estimates: $\hat{m}_t \gets \frac{m_t}{1 - \beta_1^t}$, $\hat{v}_t \gets \frac{v_t}{1 - \beta_2^t}$\;\\
    \begin{equation*}
            \textnormal{(DP-Adam)} \;\; \theta_t \gets \theta_{t-1} - \eta_t \frac{\hat{m}_t}{\sqrt{\hat{v}_t} + \gamma}
    \end{equation*}
    \begin{equation*}
            \textnormal{(DP-AdamBC)} \;\; \theta_t \gets \theta_{t-1} - \eta_t \frac{\hat{m}_t}{\sqrt{\max(\hat{v}_t-(\sigma C/L)^2, \gamma')}}
    \end{equation*}
}   
\end{algorithm}


\section{Experiments on Real Heavy-Tail Imbalanced Datasets}
\label{appendix:exp_real_data}

While linear models are convenient to demonstrate the idea mathematically, they are often limited in prediction power to fit more complex datasets. In this section we show experiment results that use deep learning models on a modified heavy-tail imbalanced image classification dataset and two next-token-prediction language datasets.

\paragraph{Datasets and Models.}

\begin{figure}[htp]
    \centering
    \includegraphics[width=0.98\textwidth]{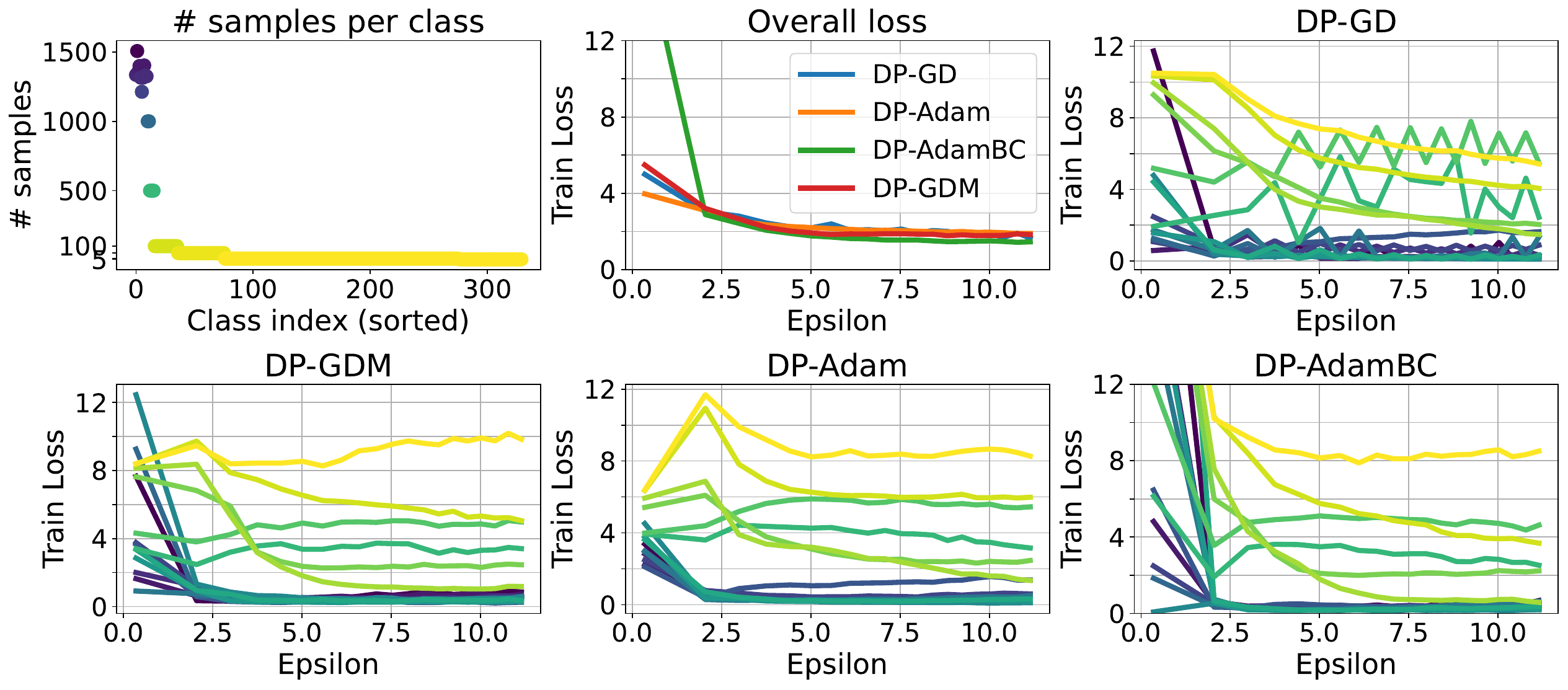}
    \caption{Results from Barcode MNIST with CNN. (a) The class distribution of the synthetic dataset. (b) The mean training loss from averaging all samples. (c)-(f) The mean training loss separated by averaging samples with the same label frequency.}
    \label{fig:barcode_mnist_c1n10_loss}
\end{figure}

\begin{figure}[htp]
    \centering
    \includegraphics[width=0.98\textwidth]{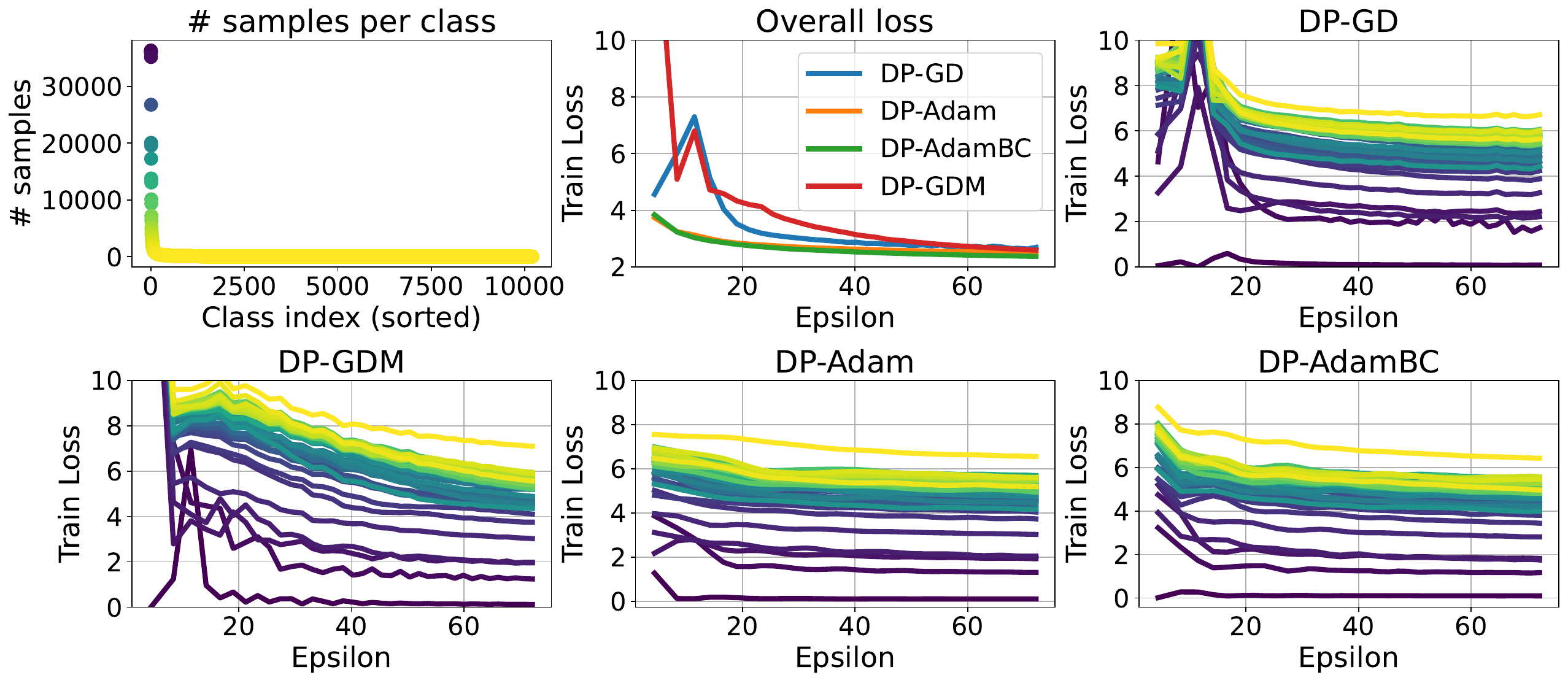}
    \caption{Results from TinyPTB with finetuned Transformer head. (a) The class distribution of the synthetic dataset. (b) The mean training loss from averaging all samples. (c)-(f) The mean training loss separated by averaging samples with the same label frequency.}
    \label{fig:ptb_c10n1_loss}
\end{figure}

\begin{figure}[htp]
    \centering
    \includegraphics[width=0.98\textwidth]{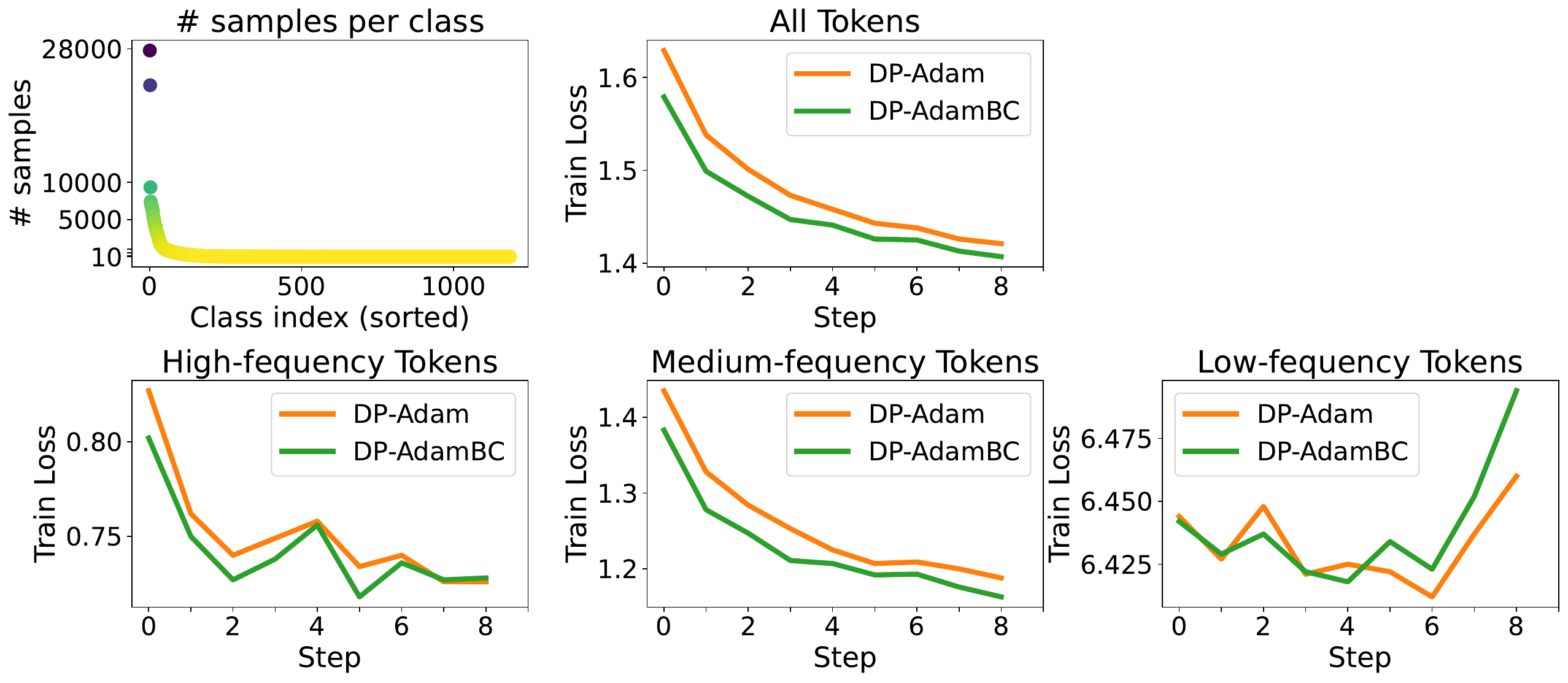}
    \caption{Results from finetuning E2E dataset with with $C=1, \sigma=0.6$. (a) The heavy-tail token frequency distribution. (b) The mean training loss from all samples. (c)-(e) The mean training loss separated by high, medium and low token frequency groups.}
    \label{fig:e2e_lora}
\end{figure}

We use the Barcode MNIST dataset which is modified from MNIST with additional classes consist of examples of original MNIST image plus a 10-bit number encoded in the corner of the image. We follow the steps of \cite{kunstner2024heavytailedclassimbalanceadam}(Appendix A2) to generate the imbalanced dataset, and we subsample examples from classes to make the labels heavy-tail imbalanced. The Barcode MNIST dataset used in our experiment has $n=23770$ images and $c=330$ numbers of classes with label distribution shown in Figure \ref{fig:barcode_mnist_c1n10_loss}(a). We train the dataset with a 2-convolution-layer CNN model.
For examples of real-world heavy-tail class imbalance datasets, we evaluate the training performance on two language datasets with next-token prediction task. We use the TinyPTB dataset \cite{kunstner2024heavytailedclassimbalanceadam}(Appendix A2) and load pre-trained weights and tokenizer from GPT2. We finetune the dataset with a Transformer head with the implementation of `GPT2LMHeadModel' from the transformers package. The token distribution is highly imbalanced and is shown in Figure \ref{fig:ptb_c10n1_loss}(a). We omit the highest frequency token (`the') with frequency $>400000$ to better visualize of the label distribution. 
In addition, we use the End-to-End (E2E) NLG challenge dataset \cite{novikova2017e2edatasetnewchallenges} with $n=42000$ training examples, which contains structured input representations in the restaurant domain that are mapped to natural language through end-to-end training. We finetune the dataset using LoRA \cite{hu2021loralowrankadaptationlarge} applied to pretrained GPT-2-medium weights and the GPT-2 tokenizer, following the implementation from \cite{yu2022differentiallyprivatefinetuninglanguage}. The token distribution exhibits heavy-tail class imbalance as shown in Figure \ref{fig:e2e_lora}(a). We use the default model and LoRA hyperparameters as \cite{yu2022differentiallyprivatefinetuninglanguage} for both optimizer setups. 
The hyperparameter tuning procedure is the same as described in section \ref{subsec:experiment_linear}. For the DP hyperparameters, we try two levels of $\sigma=\{1,10\}$ and tune a coarse set of $C=\{0.1,1,10\}$ and report the result from the best $C$.

\paragraph{Results and Discussion.}
We observe that the result on Barcode MNIST much resembles the results on the linear model example. When separating the performance by label frequency, we see that DP-GD has oscillating training loss for the medium-frequency classes, indicating these parameters potentially steps too aggressively in  high curvature regions that causes unstable performances. Except the the lowest-frequency class which all optimizers seem to suffer, we observe that DP-AdamBC has lower training loss for low-frequency classes than the other optimizers, e.g. DP-AdamBC has a training loss of 3.6 which is 0.3, 2.3, 1.2 lower than the loss of DP-GD, DP-Adam and DP-GDM respectively on the group that has 200 classes with 10 samples each. For training accuracy, we observe that all the optimizers behave similarly for the high-frequency class, with DP-GD: 97.6\%, DP-GDM: 91.7\%, DP-Adam: 97.4\%, DP-AdamBC: 96.5\%. The accuracy are smaller overall for the low-frequency classes which indicates that rare classes are indeed more difficult to optimize. However, DP-AdamBC has an advantage from estimating the curvature information more accurately: DP-AdamBC achieves 11.5\% training accuracy for the lowest-frequency group, whereas other optimizers' performance degrade more, with DP-GD: 6.6\%, DP-GDM: 4.1\%, DP-Adam: 6.3\% training accuracies respectively. 

\begin{figure}[htp]
    \centering
    \includegraphics[width=0.98\textwidth]{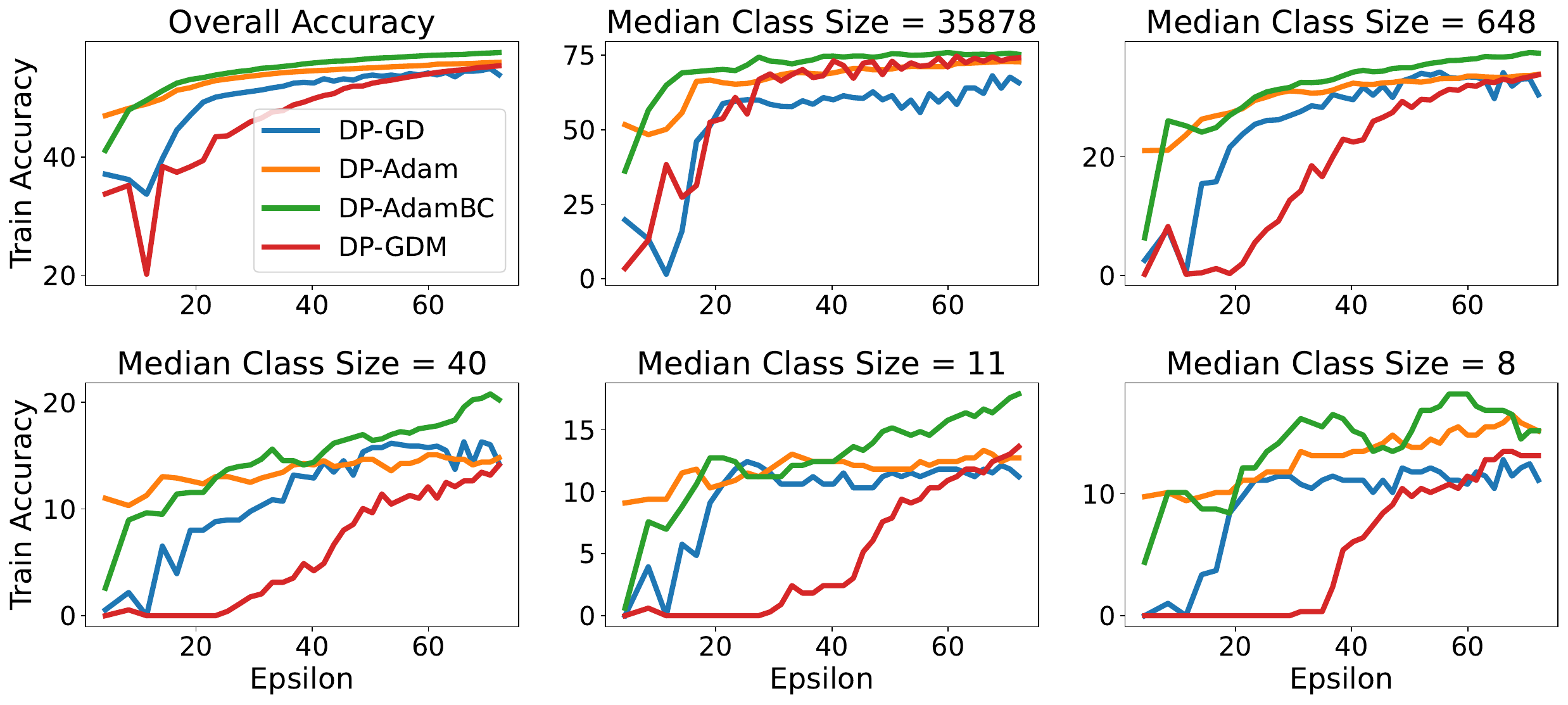}
    \caption{Results from TinyPTB with finetuned Transformer head. (a) The mean training accuracy from averaging all samples. (b)-(f) The mean training accuracy separated by averaging samples with the same label frequency.}
    \label{fig:ptb_c10n1_acc}
\end{figure}

Figure \ref{fig:ptb_c10n1_loss} and \ref{fig:ptb_c10n1_acc} shows the training loss and accuracy for the TinyPTB dataset when finetuning a transformer head from the pretrained GPT2 model. The token distribution is highly imbalanced with $\approx40\%$ of tokens of frequency $f<10$, and the top 20 high-frequency tokens spread over a wide range from $f\approx 400000$ to $5000$. Therefore we group examples such that the high-frequency tokens are grouped by coarse magnitude (group 1 with $f>400000$, group 2 with $400000> f >20000$, etc.), the medium-frequency tokens with $5000 < f < 5$ are grouped evenly into 30 groups, and all tokens with $f<5$ are combined into one group.
We observe that all optimizers could learn the highest frequency class well with low training loss, with values around 0.1 for all optimizers. For the high to medium-frequency tokens, DP-GD generally has higher training loss while the other optimizers have lower values, e.g. for the token with frequency of $\approx30000$, DP-GD has a loss of 1.7 whereas DP-Adam, DP-AdamBC and DP-GDM has 1.3, 1.2, 1.2 respectively. For the training accuracies, DP-GD have lower training accuracy than the other three optimizers in predicting medium to high-frequency tokens whereas DP-AdamBC has larger values, e.g. on the same token of frequency $\approx30000$, DP-GD, DP-Adam, DP-AdamBC and DP-GDM has 66\%, 73\%, 75\%, 74\% training accuracy respectively. For the lowest frequency class with median frequency $= 8$, all optimizers seem to behave equally poorly in terms of training accuracy though DP-GD seems to have lower training loss. In general, we see a similar but milder effect with the next-token prediction task. In addition to the more extreme imbalanced distribution and a more complex task, we suspect that finetuning from pretrained models potentially make each sample more orthogonal to each other in the embedding space, which decreases correlation between within class examples thus relieving the dependence on class frequency.

\begin{figure}[htp]
    \centering
    \includegraphics[width=0.98\textwidth]{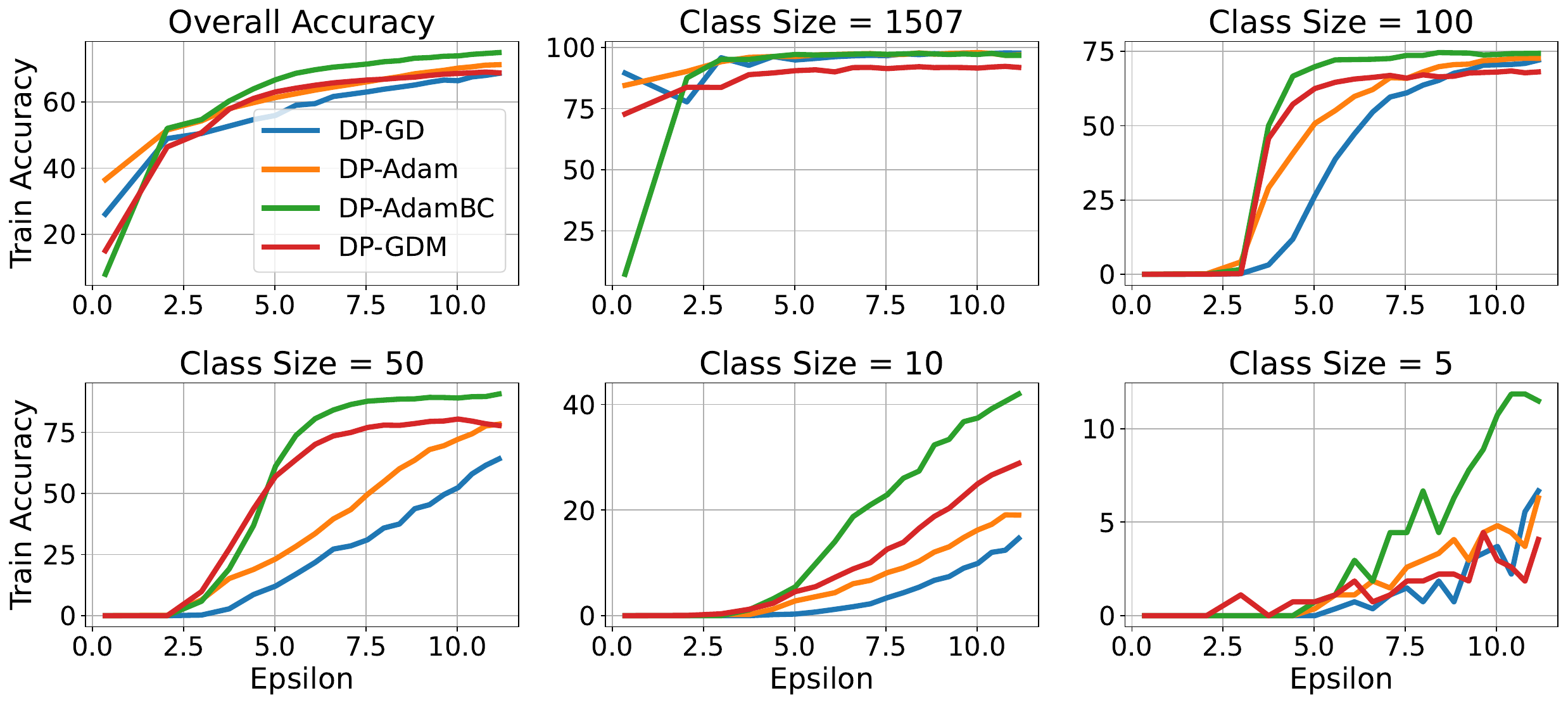}
    \caption{Results from Barcode MNIST with CNN. (a) The mean training accuracy from averaging all samples. (b)-(f) The mean training accuracy separated by averaging samples with the same label frequency.}
    \label{fig:barcode_mnist_c1n10_acc}
\end{figure}

\section{Experiments with Varying DP Hyperparameters}
\label{appendix:vary_c_sigma}
\begin{figure}[htp]
    \centering
    \includegraphics[width=0.98\textwidth]{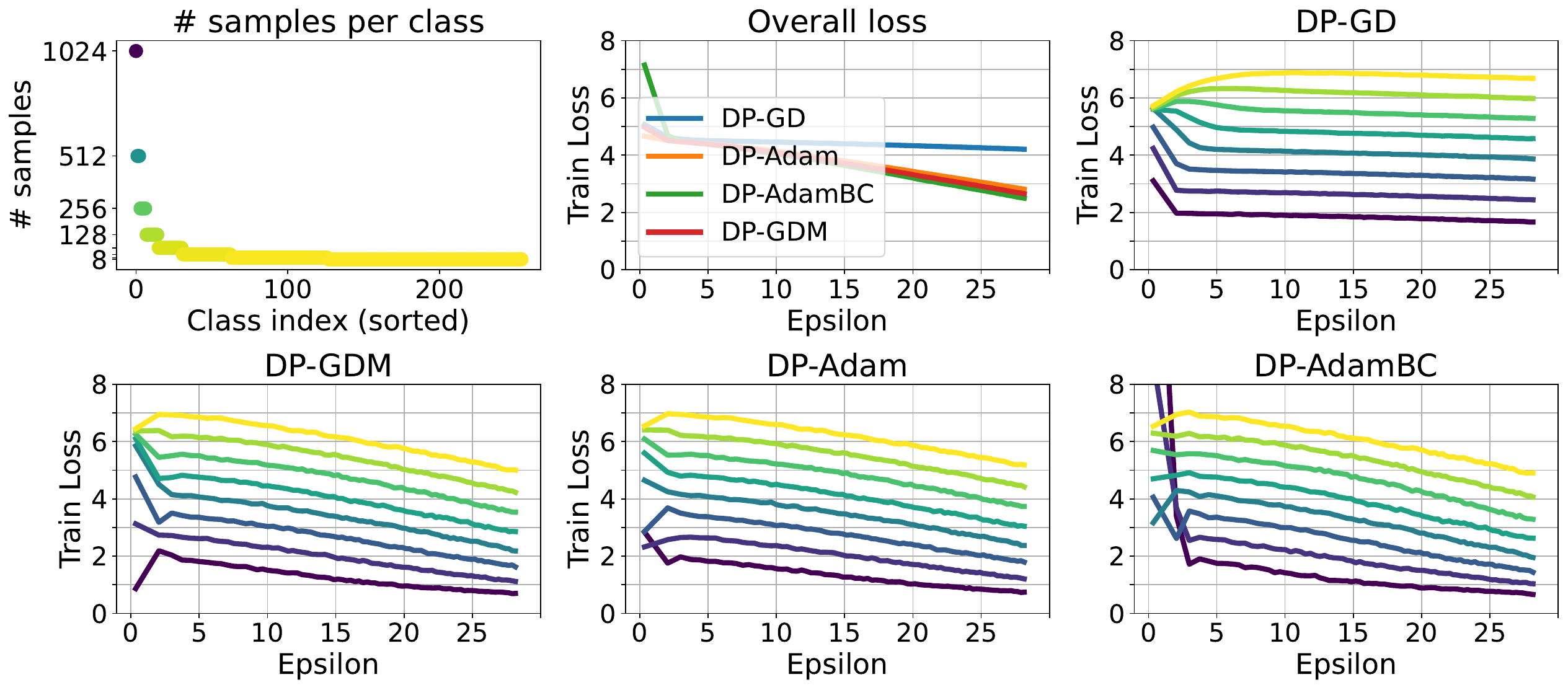}
    \caption{Results from synthetic dataset with linear model with $C=10, \sigma=10$. (a) The class distribution of the synthetic dataset. (b) The mean training loss from averaging all samples. (c)-(f) The mean training loss separated by averaging samples with the same label frequency.}
    \label{fig:c10n10_loss}
\end{figure}

\begin{figure}[htp]
    \centering
    \includegraphics[width=0.98\textwidth]{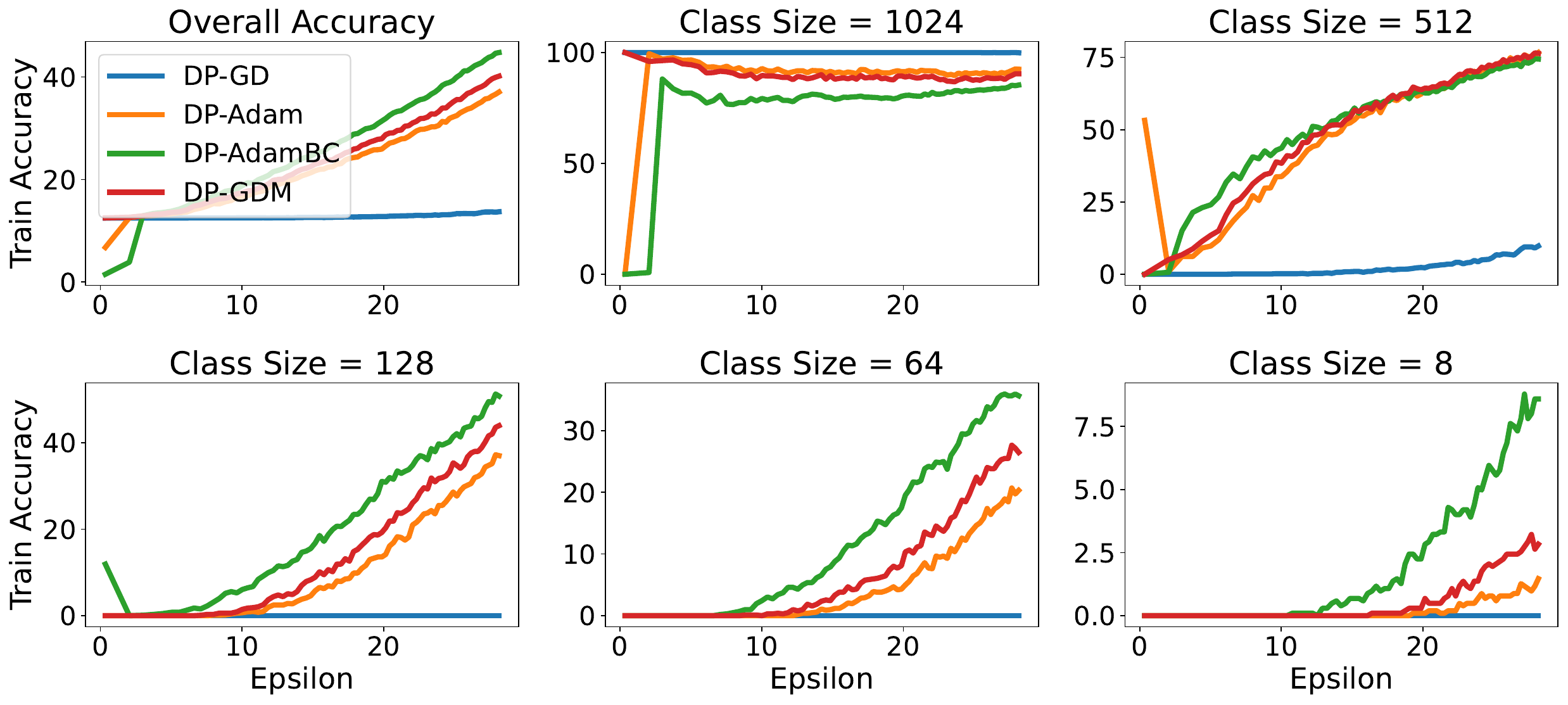}
    \caption{Results from synthetic dataset with linear model with $C=10, \sigma=10$. (a) The mean training accuracy from averaging all samples. (b)-(f) The mean training accuracy separated by averaging samples with the same label frequency.}
    \label{fig:c10n10_acc}
\end{figure}

\begin{figure}[htp]
    \centering
    \includegraphics[width=0.98\textwidth]{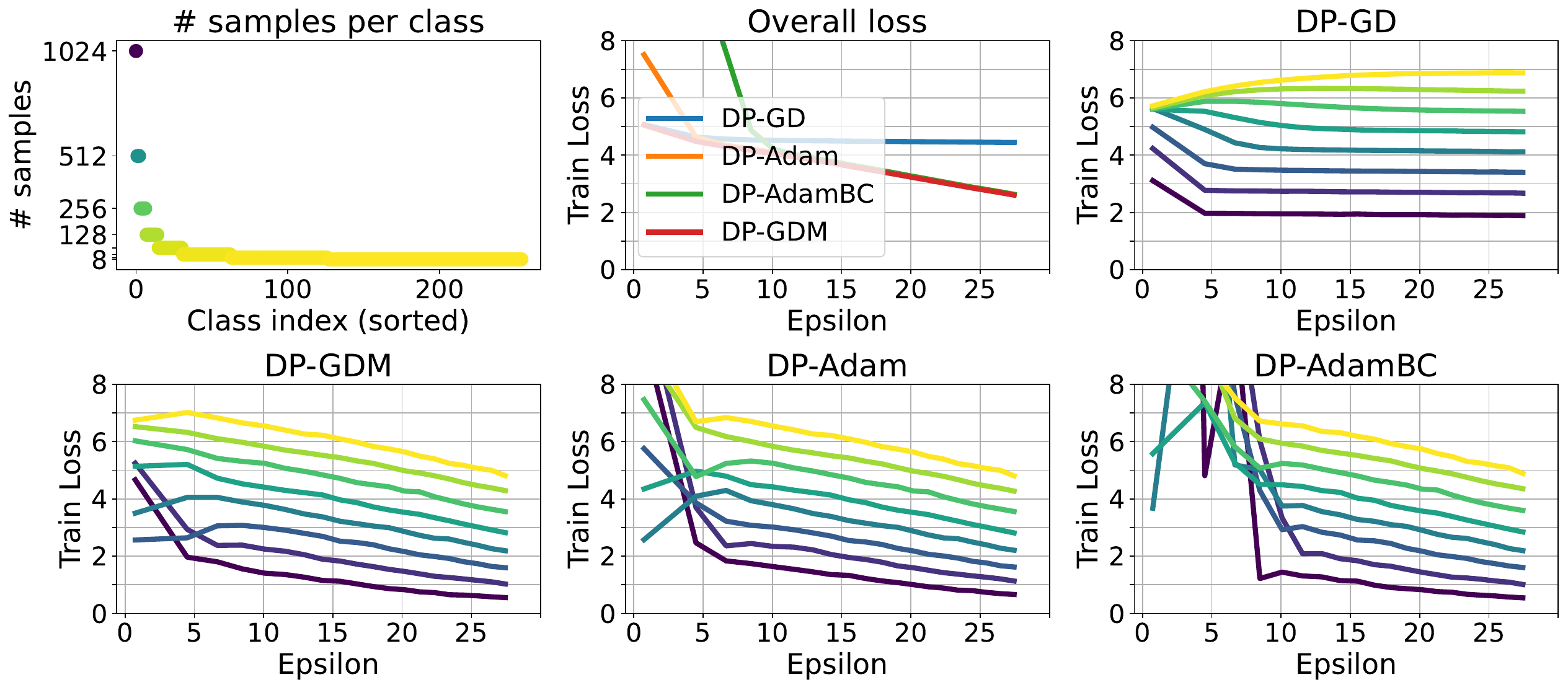}
    \caption{Results from synthetic dataset with linear model with $C=1, \sigma=5$. (a) The class distribution of the synthetic dataset. (b) The mean training loss from averaging all samples. (c)-(f) The mean training loss separated by averaging samples with the same label frequency.}
    \label{fig:c1n5_loss}
\end{figure}

\begin{figure}[htp]
    \centering
    \includegraphics[width=0.98\textwidth]{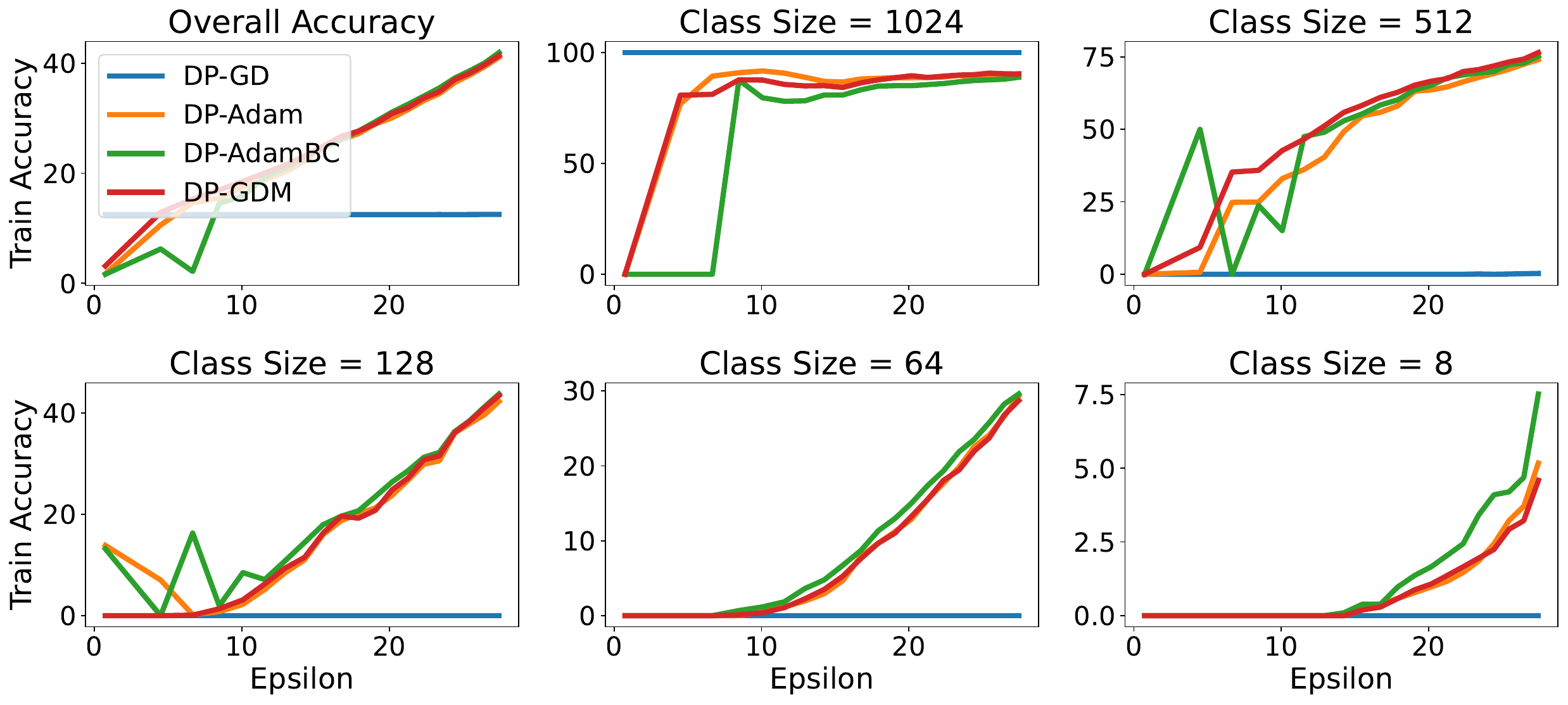}
    \caption{Results from synthetic dataset with linear model with $C=1, \sigma=5$. (a) The mean training accuracy from averaging all samples. (b)-(f) The mean training accuracy separated by averaging samples with the same label frequency.}
    \label{fig:c1n5_acc}
\end{figure}

We examine the empirical performance on the linear model with synthetic data setup when changing DP hyperparameters $C$ and $\sigma$. Figure \ref{fig:c1n10_loss} and \ref{fig:c10n10_acc} show the training loss and accuracy when $C=1, \sigma=10$. Figure \ref{fig:c10n10_loss} and \ref{fig:c10n10_acc} show the results with the same $\sigma=10$ and with a larger clipping threshold of $C=10$. Figure \ref{fig:c1n5_loss} and \ref{fig:c1n5_acc} shows the results with the same $C=10$ and with a smaller privacy parameter $\sigma=5$. We set the number of epochs such that all the experiments run until reaching the same final $\epsilon=28$. From Figure \ref{fig:c10n10_loss} and \ref{fig:c10n10_acc} we observe that changing the clipping threshold shows approximately the same trends as in Figure \ref{fig:c1n10_loss} and \ref{fig:c10n10_acc}, which supports the claim in \S\ref{subsec:clipping-ill-condition} that clipping per-sample gradients to smaller scales does not relieve the ill-condition in DP-GD and requires more accurate estimates of curvature with DP bias removed to see better performance in learning the low-frequency classes. Comparing Figure \ref{fig:c1n5_loss} and \ref{fig:c1n5_acc} to the others, we observe that DP-GD suffer from learning lower-frequency classes due to the ill-condition. However, when the DP noise variance $(\sigma C/n)^2$ is smaller, DP-Adam could learn the low-frequency classes similarly as DP-AdamBC, potentially from estimating the second moments more accurately. 

\end{document}